\documentclass{article}
\usepackage{ijcai19}

\usepackage{times}
\usepackage{soul}
\usepackage{url}
\usepackage{hyperref}
\usepackage[utf8]{inputenc}
\usepackage[small]{caption}
\usepackage{graphicx}
\usepackage{subcaption}
\usepackage{amsmath}
\usepackage{booktabs}
\usepackage{algorithm}
\usepackage{algorithmic}
\usepackage{helvet}
\usepackage{courier}
\usepackage{amsfonts}
\usepackage{booktabs}
\usepackage{threeparttable}
\usepackage{multirow}
\title{Cross-filter compression for CNN inference acceleration}

\author{
Fuyuan Lyu$^{1,2}$ \footnote{Work is done while visiting NTU PDCL Lab}
\and
Shien Zhu$^1$
\and
Weichen Liu$^1$\footnote{Contact Author}
\affiliations
$^1$Nanyang Technological Univerisity\\
$^2$McGill Univerisity
\emails
fuyuan.lyu@mail.mcgill.ca,
shien001@e.ntu.edu.sg,
liu@ntu.edu.sg
}

\begin{document}

\maketitle

\begin{abstract}

Despite the great success of convolutional neural networks on various applications, the rigid computational resource limits their wide usage on edge devices. To address this issue, many research works have been done upon uniform quantization and compact network design, which are commonly used compression methods to reduce the computational requirement. We observe that small kernels, widely adopted in compact network design, can reduce the acceleration upper bound of uniform quantization, which leads to a conflict that the acceleration ratio achieved by both methods may be smaller than that of uniform quantization alone. This phenomenon is related to the filter-wise limitation that one scaling factor corresponds to one filter in uniform quantization. In this paper, we propose a new quantization method, called cross-filter compression, to break the filter-wise limitation and boost the speedup effect when combining two methods. It can provide  $122\times$ speed up in convolution operations and $\sim 32 \times$ model size reduction when integrated upon XNOR-Net. Our method can be implemented on different filter-wise uniform quantization methods with scaling factor. It is evaluated on CIFAR-10 and ImageNet dataset with widely used network structures, such as ResNet and VGG, and -0.3\% to 0.4\% accuracy loss is witnessed compared to original filter-wise quantification methods. 

\end{abstract}

\section{Introduction}

Since the introduction of deep convolution neural networks\cite{krizhevsky2012imagenet}, significant achievements have been made in terms of solving complicated tasks like image classification\cite{simonyan2014very}, object detection\cite{redmon2016you} and object segmentation\cite{he2017mask}.

However, these powerful networks require millions of full-precision operations for one single input, thus increasing the inference time upon different hardware. The inference time for one single ImageNet dataset input\cite{imagenet_cvpr09} can be up to 650ms on Samsung S5 and nearly 1.2s on Moto E\cite{li2018deeprebirth}. Hence, deploying a network for real-time application tends to have rigid hardware requirement (normally on GPU). Such high requirement makes these neural networks difficult to be deployed on edge devices like personal mobile phones or wearable devices, which only have limited computation power and energy. 

Many research works have been done to address this hard-to-deploy issue. Basically these methods can be classified into two categories\cite{sze2017efficient}. First is to reduce precision, including uniform quantization\cite{rastegari2016xnor} \cite{courbariaux2016binarized}\cite{courbariaux2015binaryconnect}. Among them XNOR\cite{rastegari2016xnor} is a representative approach. Second is to reduce the number of operations and model size, such as compact network design\cite{szegedy2016rethinking}. However, these methods alone do not meet the requirement for real-time application on edge devices.

In order to do inference on edge devices for real-time applications, we intend to combine these two kinds of methods to reach the requirement. Compact network design and uniform quantization are selected as the representatives. However, when combining these two methods, we witness that the usage of compact network structure, especially the small kernels, will constrain the acceleration ratio to a certain extent. \textbf{And more over, there even exists such possibility that uniform-quantized compact networks may require more computational power and memory occupation compared to classic networks with uniform quantization alone.}

This is because during the uniform quantization process, filter-wise scaling factors are normally adopted to scale up the tensor to match the original tensor, defined as filter-wise limitation. Above filter-wise limitation results to the fact that the number of scaling factors increases linearly with the channel of the tensors. And the number of scaling factors also impacts the number of full-precision multiplication operations, which cost significant more clock cycles compared to other commonly used operations\cite{sze2017efficient}. Meanwhile, with the popularity of small kernels, such as $3 \times 3$ and $1 \times 1$ kernels, and deep channels\cite{szegedy2016rethinking}\cite{chollet2017xception}, it can be foreseen that the acceleration ratio achieved by uniform quantization of compact neural networks is decreasing. 

In this paper, for the first time, we address the conflict between uniform quantization and compact network design, and propose a cross-filter compression method, which can bridges such gap and boost the speedup effect when combining them. The key idea of our method is to share numerical similar scaling factors cross filters to reduce the number of full-precision operations, and further, to reduce the inference time. Our cross-filter compression method is suitable for compressing and accelerating small filters and can be integrated to quantization methods with scaling factors. We implement our cross-filter compression methods on XNOR and Binary weight quantization\cite{rastegari2016xnor}. We are able to achieve up to $122\times$ speedup and $\sim 32 \times$ memory savings with less than 0.3\% accuracy lost in worst case and up to 0.4\% accuracy gain in best case compared to corresponding single-filter compression methods.

\section{Related Work}
\subsection{Uniform Quantization}
\subsubsection{Quantizing weight}
Weight Quantification is an essential and efficient method to compress and accelerate deep convolutional neural networks. In short, it uses fewer bits to represent weights. Binary weight, which use +1 or -1 to represent weights, is widely used. \cite{courbariaux2015binaryconnect} proposed deterministic and stochastic functions to quantify weights to binary numbers.  \cite{rastegari2016xnor} proposed a similar network but with a filter-wise scaling factor. Another idea is to use ternary weight, or +1, 0, -1  to represent weights. \cite{1605.04711} first proposed this idea and \cite{zhu2016trained} introduced two scaling factors for positive and negative weights respectively.

\subsubsection{Quantizing both weight and activation }
Simply quantizing weights reduce the model size greatly, but it still requires many computation resources during inference stage. Hence, some researchers proposed activation quantization. \cite{courbariaux2016binarized} quantized both weight and activation with 1-bit. Similarly, \cite{rastegari2016xnor} also proposed XNOR-Net but with filter-wise scaling factors for both. \cite{li2017performance} introduced high-order residual quantification to leverage accuracy lost and speed up. \cite{zhou2016dorefa} presented a method to train neural networks with low bit-width numbers of both weights and activation. \cite{wan2018tbn} proposed a method to combine binary weight and ternary activation.

\subsection{Compact Network Design}
Apart from reducing the complexity of existing methods, some researchers intended to design more efficient operations to build compact networks.\cite{szegedy2016rethinking} demonstrated that small filter is more efficient than large filter. \cite{howard2017mobilenets} and \cite{chollet2017xception} invented depth-wise separate convolution, a more efficient convolution structure. \cite{hu2017squeeze} proposed a squeeze-and-extraction operation on channel level, which increased the accuracy with nearly no computational cost increase. \cite{ma2018shufflenet} proposed a shuffle operation to increase the network capability with no extra layer.

\section{Dilemma when combining uniform quantization and compact network design} \label{dilemma section}


In this section, we will mathematically discuss the dilemma between uniform quantization and compact network design when combining these two methods. We notice that the computational cost of quantized compact network may be even higher than classic network with only uniform quantization under certain circumstance.

Generally speaking, we represent one convolutional layer with a triplet $\langle \mathbf{I}, \mathbf{W}, *\rangle$. $\mathbf{I} \in \mathbb{R}^{i_h \times i_w \times i_c}$ is the input tensor and $\mathbf{W} \in \mathbb{R}^{k_h \times k_w \times i_c \times o_c}$ is the filter. Here $i_h$, $i_w$, $i_c$, $k_h$, $k_w$ and $o_c$ represents input heights, input widths, input channels, kernel heights, widths and output channels respectively. By applying convolution operation\cite{dumoulin2016guide}, we can get the output $\mathbf{O} \in \mathbb{R}^{o_h \times o_w \times o_c}$, where $o_h$, $o_w$ represent output heights and widths. We define output shape $N_o = o_ho_w$ for the sake of simplicity.

Suppose we have a filter $\mathbf{W}_1 \in \mathbb{R}^{k_{h1} \times k_{w1} \times i_c \times o_c}$, where $k_{h1} = k_{w1} = 5$. According to \cite{szegedy2016rethinking}, one $5 \times 5$ kernel can be replaced by two $3 \times 3$ kernel with slight accuracy improvement. Hence, the filter $\mathbf{W}_1$ can be replaced with two filter  $\mathbf{W}_2 \in \mathbb{R}^{k_{h2} \times k_{w2} \times i_c \times o^{'}_c}$ and $\mathbf{W}_3 \in \mathbb{R}^{k_{h2} \times k_{w2} \times o_c^{'} \times o_c}$, where $k_{h2} = k_{w2} = 3$.  W.l.o.g. we assume that $o_c^{'}=o_c = i_c$ and $N_{o1}=N_{o2}=N_{o3}=N_o$, where $N_{o1}$,$N_{o2}$ and $N_{o3}$ represent the corresponding output shape. Based on these information, we can compute the parameter memory size and floating point operations, as shown in Table \ref{dilemmatable}. We can see that using two $3\times3$ kernels outperform one $5\times5$ kernel in both FLOPs and parameter size.

Moreover, by applying binary weight quantization upon these filters, we get  $\mathbf{B}_1 \in {\{+1,-1\}}^{5 \times 5 \times i_c \times i_c}$,  $\alpha_1 \in \mathbb{R}^{i_c}$ for $5\times5$ kernels and $\mathbf{B}_2, \mathbf{B}_3 \in {\{+1,-1\}}^{3 \times 3 \times i_c \times i_c}$, $\alpha_2, \alpha_3 \in \mathbb{R}^{i_c}$ for two $3\times3$ kernels. The parameter memory and FLOPs are also listed in Table \ref{dilemmatable}. The computation of original layer requires $25i_c^2N_o$ binary operations and $i_cN_o$ full-precision multiplication-addition-cumulative (MAC) operations. Meanwhile, the computation of compact layer requires $18i^2_cN_o$ binary operations and $2i_cN_o$ full-precision MAC operations. We observe that after quantization , the compact network requires even more full-precision MAC operations compared to the original neural network.

\begin{table}[htb]
\caption{An example about the dilemma}
\label{dilemmatable}
\centering
\begin{tabular}{c|ccc}
\hline
cases & &$5\times5$ kernel & $2\times3\times3$ kernel \\
\hline
\multirow{2}{*}{\rotatebox{90}{FP}} & FLOPs & $25i^2_cN_o$ & $18i_c^2N_o$  \\
& Params & $100i_c^2$ & $72i_c^2$ \\
\hline
\multirow{4}{*}{\rotatebox{90}{QWQA}}
& Binary ops & $25i^2_cN_o$ & $18i^2_cN_o$ \\
& FLOPs & $i_cN_o$ & $2i_cN_o$ \\
& Binary Params & $\lceil \frac{25i_c^2}{8} \rceil$& $2\lceil \frac{9i_c^2}{8} \rceil$ \\
& Float Params & $4i_c$ & $8i_c$ \\
\hline

\end{tabular}
\begin{tablenotes}
\item[-] \small In this table, we compare the floating point operations (FLOPs), model size between $5\times5$ kernel case and $2\times3\times3$ kernel case before and after binary quantization.
\end{tablenotes}
\end{table}

Suppose the modern hardware (e.g. CPU, GPU, ASIC, FPGA) can perform $L$-bits binary operation in one clock cycle (typically, L=64) and the ratio between a multiply-accumulate operation and performing L-bits binary operation is $\gamma$ like \cite{wan2018tbn}. The quantized $5\times5$ kernel can be computed no slower than two $3\times3$ kernels when:
\begin{equation} \label{dilemma}
\small {\frac{1}{L}25i_c^2N_o + \gamma i_cN_o \leq \frac{1}{L}18i_c^2N_o + 2\gamma i_cN_o}
\end{equation}
which can be  simplified as:
\begin{equation}
\small{ i_c \geq \frac{\gamma}{7}L}
\end{equation}
Here we adopt $\gamma = 1.91$ as \cite{wan2018tbn}. If $L=64$, the boundary is 18. But with the increase of instruction length, such as AVX-512 where $L=512$, the boundary becomes 140, which is not an ignorable number. This number indicates that in certain cases, the computation time of quantized compact network is even longer than the simply quantized network, which is the key dilemma we are facing.

\section{Cross-filter Compression Methodology}
\begin{figure}[htb]

\centering
\includegraphics[width=0.4\textwidth]{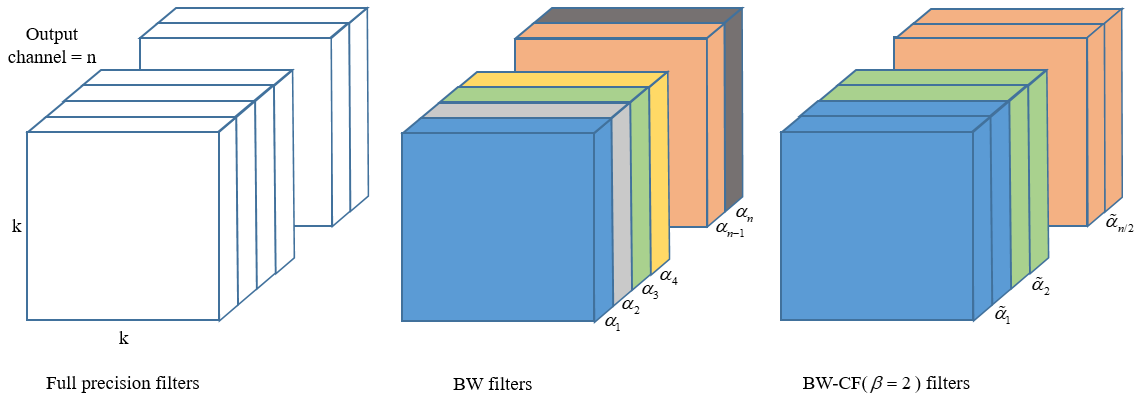}
\caption{\small This is a figure demo for our proposed cross-filter compression methods. The left column represents two full-precision filters. The middle column is the compressed result based on Binary-Weight-Network(BW). Each filter is represented by a different scaling factor. The right column is our cross filter compression result with filter compression factor $\beta=2$. Same color represents that they share the same scaling factor.}
\label{fig:illus2}
\end{figure}

We propose a cross-filter compression method which is able to compress $\beta$ spatial-adjunct filters using one scaling factor. The number $\beta$ is a hyper-parameter predefined by users. The illustration figure is shown in Fig.\ref{fig:illus2}. We propose the case implemented upon Binary Weight Network\cite{rastegari2016xnor}, in which each individual weight is represented with 1-bit. We follow the same quantization rule. However, several neighboring filters share the same scaling factor, instead of adopting filter-wise scaling factors. This can reduce the model size, increase the parallelism and accelerate the training and inference period.

\subsection{Optimal result for cross-filter compression integrated upon Binary Weight Network}
The quantification problem is to find a pair $(\mathbf{A},\mathbf{B})$ that best represent a slice tensor $\mathbf{W}$ from an original weight tensor. Here $\mathbf{A}$ represents the scaling matrix, while $\mathbf{B}$ represents the restored matrix after quantification. If the target restored matrix is binary like\cite{rastegari2016xnor}, then the $\mathbf{B}$ is composed of +1 and -1. And the scaling matrix $\mathbf{A}$ can be further written as $\mathbf{A} = \alpha * \mathbf{I}$, and $\alpha$ is called the scaling factor, as it is in \cite{courbariaux2016binarized}, and $\mathbf{I}$ is unit matrix. Here $\alpha \in \mathbb{R}$, $\mathbf{B} \in {\{+1,-1\}}^{k_h \times k_w \times i_c \times \beta}$, $\mathbf{W} \in {\mathbb{R}}^{k_h \times k_w \times i_c \times \beta}$ , $\beta$ is the number of filters chosen to compress as one, $i_c$ and $o_c$ indicate the input channel and output channel, $k_h$ and $k_w$ refer to kernel heights and kernel widths. Assume that the tensor is normalized previously. In that case, the optimization problem can be rewritten as the following:

\begin{equation}
\begin{array}{lr}
             J(\mathbf{B},\alpha)=||\mathbf{W}-\alpha \cdot \mathbf{B}||^2  &  \\
            \mathbf{B}^{*},{\alpha}^{*} = {argmin}_{\mathbf{B},\alpha} J(\mathbf{B},\alpha) &  
\end{array}
\end{equation}

After expansion, we get:
\begin{equation}
J(\mathbf{B},\alpha) = {\alpha}^2\mathbf{B}^T\mathbf{B} - 2\alpha \mathbf{W}^T\mathbf{B} + \mathbf{W}^T\mathbf{W}
\end{equation}
Here we first set $\alpha$ as constant, and take a look at $B$. The first and third item would be constant. And we get:
\begin{equation}
\mathbf{B}^* = argmax_\mathbf{B}\mathbf{W}^T\mathbf{B}
\end{equation}
It's easy to know that:
\begin{equation} \label{eq:B1}
\mathbf{B}^* = sign(\mathbf{W})
\end{equation}
And we suppose the derivation of $J(\mathbf{B},\alpha)$ over $\alpha$ to $0$ so we can get $\alpha^*$.
\begin{equation}
2\alpha \mathbf{B}^T\mathbf{B} - 2 \mathbf{W}^T\mathbf{B} = 0
\end{equation}
\begin{multline} \label{eq:alpha1}
\alpha^* = \mathbf{W}^T\mathbf{B} / \mathbf{B}^T\mathbf{B} = \mathbf{W}^T  sign(\mathbf{W}) / (\beta k_h k_w i_c ) \\
	= \sum |\mathbf{W}_i|/n = \frac{1}{n} \lVert \mathbf{W} \rVert_{l1}
\end{multline}

\begin{figure}[htb]
\centering
\includegraphics[width=0.4\textwidth]{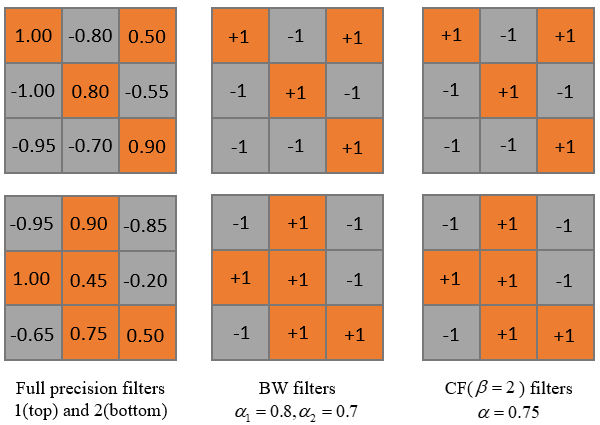}
\caption{\small This is a figure demo to illustrate our proposed cross-filter compression methods. The left column is two 3x3 full-precision filters. The middle column is the compressed result based on Binary-Weight-Network(BW) with two different scaling factors $\alpha_1$ and $\alpha_2$. The right column is our cross filter compression result with filter compression factor $\beta=2$. Both filters share the same scaling factor $\alpha$.}
\label{fig:illus}
\end{figure}

So in our binary experiment, we select $\alpha$ as the absolute mean of the whole tensor slice and $\mathbf{B}$ as the signs of the corresponding numbers in the tensor slice. Fig.\ref{fig:illus} illustrates the selection of scaling factor $\alpha$ given $\beta=2$.

\subsection{Optimal result for cross-filter compression integrated upon XNOR}

Once the quantization of activation is taken into consideration, the approximation target change to the dot product between $\mathbf{H}, \mathbf{B} \in{\{+1,-1\}}^n $ . In this case, the optimization problem become the following:
\begin{equation} \label{eq:1}
    \alpha ^ { * } , \mathbf { B } ^ { * } , \beta ^ { * } , \mathbf { H } * = \underset { \alpha , \mathbf { B } , \beta , \mathbf { H } } { \operatorname { argmin } } \| \mathbf { X } \odot \mathbf { W } - \beta \alpha \mathbf { H } \odot \mathbf { B } \|
\end{equation}
where $\odot$ indicates element-wise product. We define $\mathbf { Y } \in \mathbb { R } ^ { n }$ such that $\mathbf { Y } _ { i } = \mathbf { X } _ { i } \mathbf { W } _ { i }$, $\mathbf { C } \in \{ + 1 , - 1 \} ^ { n }$ such that $\mathbf { C } _ { i } = \mathbf { H } _ { i } \mathbf { B } _ { i }$ and $\gamma \in \mathbb { R } ^ { + }$ such that $\gamma = \beta \alpha$. Hence, the equation \ref{eq:1} can be rewritten as:
\begin{equation}
    \gamma ^ { * } , \mathbf { C } ^ { * } = \underset { \gamma , \mathbf { C } } { \operatorname { argmin } } \| \mathbf { Y } - \gamma \mathbf { C } \|
\end{equation}
The optimal solution of this can be written as:
\begin{equation}
    \mathbf { C } ^ { * } = \operatorname { sign } ( \mathbf { Y } ) = \operatorname { sign } ( \mathbf { X } ) \odot \operatorname { sign } ( \mathbf { W } ) = \mathbf { H } ^ { * } \odot \mathbf { B } ^ { * }
\end{equation}
\begin{multline}
    \small \gamma ^ { * } = \frac { \sum \left| \mathbf { Y } _ { i } \right| } { n } = \frac { \sum \left| \mathbf { X } _ { i } \right| \left| \mathbf { W } _ { i } \right| } { n } \approx \\
   \small \left( \frac { 1 } { n } \| \mathbf { X } \| _ { \ell 1 } \right) \left( \frac { 1 } { n } \| \mathbf { W } \| _ { \ell 1 } \right) = \beta ^ { * } \alpha ^ { * }
\end{multline}

\subsection{Training CNN with Cross-filter compression}

So far the quantization methods based on Binary-Weight-Net and XNOR-Net have been discussed. Each iteration of training CNN involves forward pass, backward pass and parameter update. We only quantize weights or activation during the first and second stages. Algorithm \ref{Algorithm:1} demonstrates our procedure about training a CNN with cross-filter compression method. We first quantize weight $\mathcal{W}^t$ according to the corresponding methods mention in above sections. Then we call forward propagation, activation compression method is adopted accordingly during this stage. What's more, in the backward pass, we adopt the same strategy used in \cite{rastegari2016xnor} to calculate gradients. Finally, parameters and learning rate gets updated by an update rule. We adopt the same approach as \cite{courbariaux2016binarized} to compute the gradients of sign function $\frac{\partial sign}{\partial r} = r 1_{\lvert r \rvert \leq 1}$. The gradients are computed as:
\begin{equation}
\frac{\partial C}{\partial \mathbf{W}_i} = \frac{\partial C}{\partial (\alpha \mathbf{B})} \frac{\partial (\alpha \mathbf{B})}{\partial \mathbf{W}_i} = \frac{\partial L}{\partial (\alpha \mathbf{B})} (\frac{1}{n} + \alpha \frac{\partial sign}{\partial \mathbf{W}_i}) 
\end{equation}

\begin{algorithm}
 \caption{Training CNN with cross-filter compression}
 \label{Algorithm:1}
 \begin{algorithmic} [1]
 \REQUIRE A minibatch of inputs and targets ($\mathbf{I},\mathbf{Y}$), loss function $C ( \mathbf { Y } , \hat { \mathbf { Y } } )$, weight $\mathcal{W}^t$ and learning rate $\mathcal{\eta}^t$
 \ENSURE Updated weight $\mathcal{W}^{t+1}$ and learning rate $\mathcal{\eta}^{t+1}$
 \STATE Quantizing weight filters with cross filter compression:
 \FOR {each layer in the network}
 \FOR {every $\beta$ filters in the layer}
 \STATE Calculate quantized weight slice $\mathcal{\hat{W}}_i^{t}$
 \ENDFOR
 \STATE Concatenate $\{ \mathcal{\hat{W}}_1^{t}, ... ,\mathcal{\hat{W}}_{n(\beta)}^{t}  \}$ as  $\mathcal{\hat{W}}^{t}$
 \ENDFOR
 \STATE $\hat { \mathbf { Y }}$= \textbf{Forward}$(\mathbf{I},\mathcal{\hat{W}}^{t})$ \COMMENT{Standard Forward propagation except that activation is quantized accordingly and convolutions are computed on quantized weight}
 \STATE $\frac { \partial C } { \partial \mathcal{\hat{W}}^{t} } = $ \textbf{BackWard}$(\frac { \partial C } { \partial \hat { \mathbf { Y } } } , \mathcal{\hat{W}}^{t})$ \COMMENT{Standard Backward propagation except that quantized activation is used and gradients are computed upon quantized weight}
 \STATE $\mathcal{W}^{t+1} = $ \textbf{UpdateParameters} $(\mathcal{W}^t, \frac { \partial C } { \partial \mathcal{\hat{W}}^{t}},  \mathcal{\eta}^t)$
\STATE $\mathcal{\eta}^{t+1} = $ \textbf{UpdateLearningRate} $(\mathcal{\eta}^{t},t)$
 \end{algorithmic}
 \end{algorithm}

\subsection{Finetuning from pretrained Binary Weight or XNOR Net model}
In this section, we will further prove that our cross-filter compression methods can be finetuned on pre-trained Binary Weight or XNOR net models (these two are the same regarding model storage), which is very useful for implementing them on existing models.

Here we suppose that we compress $\beta$ filters with one scaling factor. The weight of these filters are $W_1, W_2, ... , W_\beta$. Here $W_i \in \mathbb{R}^{k_h \times k_w \times i_c}$. We concatenate these filters as one and get $W$, where $W\in \mathbb{R}^{k_h \times k_w \times i_c \times \beta}$. Based on \cite{rastegari2016xnor}, we can obtain that the filter-wise optimal compression result:

\begin{equation}
\left\{
             \begin{array}{lr}
             \mathbf{B}_i^* = sign(\mathbf{W}_i)  &  \\
             \alpha_i^* = \sum \lvert \mathbf{W}_i \rvert /n &  
             \end{array}
\right.
\end{equation}

From equation \ref{eq:B1} and \ref{eq:alpha1}, we can obtain that:
\begin{multline}
    \mathbf{B}^{*} = sign(\mathbf{W}) = sign(\mathbf{W}_1,\mathbf{W}_2,...,\mathbf{W}_\beta) \\
    = (sign(\mathbf{W}_1),sign(\mathbf{W}_2),..,sign(\mathbf{W}_\beta)) \\
    = (\mathbf{B}_1^*,\mathbf{B}_2^*,...,\mathbf{B}_\beta^*) 
\end{multline}
\begin{multline}
    \alpha^{*} = \frac{1}{n} \lVert \mathbf{W} \rVert_{l1} = \sum_{i=1 ... \beta}\sum \lvert \mathbf{W}_i \rvert / ( k_h k_w i_c \beta )  \\
    = \frac 1 \beta \sum_{i=1 ... \beta}\sum \lvert \mathbf{W}_i \rvert / ( k_h k_w i_c) = \frac 1 \beta  \sum_{i=1 ... \beta} \alpha_{i}^*
\end{multline}
So far, we have proven that the optimal result $\alpha^*$ and $\mathbf{B}^*$ for our cross-filter compression can be calculated directly from the compressed pre-trained result. Hence, our cross-filter compression method can be implemented from pre-trained binary weight models without any necessity to access the full-precision model.

\section{Experiment}

\subsection{Theoretical Efficiency Analysis}
In this section, we will theoretically analyze the efficiency of our cross-filter compression methods. Notice that a standard full-precision convolution operations requires $o_c N_o N_k$ multiplication operations., where $N_k = k_h k_w i_c$ and $N_o = o_h o_w$. Our XNOR-based cross-filter compression methods approximate a standard convolution with $o_c N_o N_k$ binary operations and  $o_c N_o / \beta$ full-precision operations. Suppose the modern hardware (e.g. CPU, GPU, ASIC, FPGA) can perform $L$-bits binary operation in one clock cycle and the ratio between a multiply-accumulate operation and performing one L-bits binary operation is $\gamma$ like \cite{wan2018tbn}. Hence, the speed up ratio be
\begin{equation}
Speedup = \frac{\gamma o_c N_o N_k}{\gamma o_c N_o / \beta + \frac{1}{L} o_c N_o N_k } = \frac{1}{\frac{1}{\beta N_k} + \frac{1}{\gamma L} }
\end{equation}

\begin{figure}[htb]
\begin{subfigure}{.23\textwidth}
  \centering
  \includegraphics[width=\textwidth]{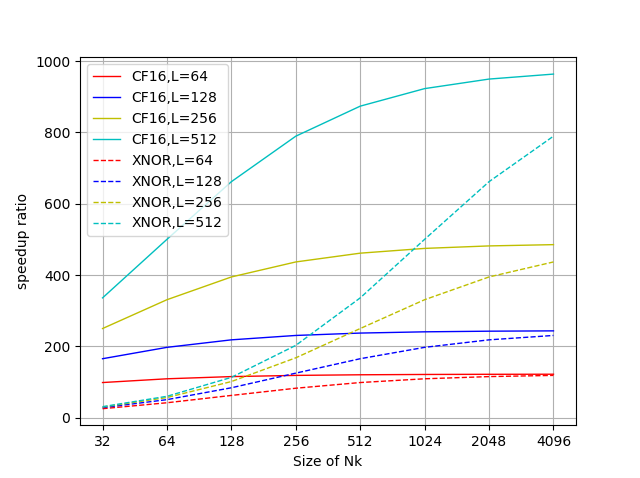}
  \caption{}
  \label{fig:sfig2-L}
\end{subfigure}%
\begin{subfigure}{.23\textwidth}
  \centering
  \includegraphics[width=\textwidth, height=0.8\textwidth]{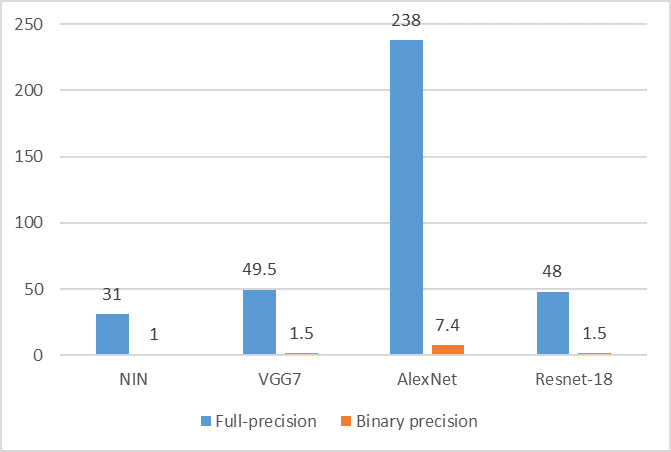}
  \caption{}
  \label{fig:sfig2-size}
\end{subfigure}
\caption{\ref{fig:sfig2-L}: The relationship between speedup ratio and $N_k$ under different L and compression methods. \ref{fig:sfig2-size}: The efficiency of binary quantization in terms of memory.}
\label{fig:fig1}
\end{figure}

Here, we adopt the same network configuration as \cite{wan2018tbn} by setting $N_k=2304$ , $\gamma = 1.91$ and $L=64$. Our cross filter compression method can achieve up to $122 \times$ speed up. However, with the going parallelism in hardware design, such as Intel AVX-512 can perform 512 bits binary operation in one clock cycle, and with the decreasing of kernel size, our cross-filter compression methods can achieve much higher speedup ratio compare to corresponding filter-wise compression methods. If we set $L=512$ and $N_k = 1 \times 1 \times 256 = 256$, then our XNOR-CF16 can achieve $789 \times$ speedup, nearly four time than XNOR's $203\times$ speedup. The relationship between speedup ratio and other variables is shown in Fig \ref{fig:sfig2-L}.  It can be observed that our XNOR-CF network outperforms its corresponding layer-wise compression method in all cases. However, given the fact that we exclude the data movement and memory access time, the actual result may decrease.

Our cross-filter compression method can also significantly reduce the model size by up to $\sim 32 \times$. Fig \ref{fig:sfig2-size} shows the model size comparison between full-precision model and binary precision model.

We also compare our result with other state-of-the-art methods in Table \ref{analyzetable}.  We propose two series of networks: binary-weight-network based cross filter compression (BW-CF) and XNOR-network based cross filter compression (XNOR-CF). Compared to other weight quantization method, our BW-CF network can speedup convolution operation $\sim 2$ times.  As for weight and activation quantization method, our XNOR-CF network outperforms other state-of-the-art methods. However, given the fact that we exclude the data movement and memory access time, the result may not be accurate.

\begin{table*}[htbp]
\caption{Theoretical Analyze Result}
\label{analyzetable}
\centering
\resizebox{0.8\textwidth}{!}{
\begin{tabular}{c|cccccccc}
\hline
\hline
& Methods &  Inputs & Weights & MACs & Binary Operations & Operations & Speedup \\
\hline
\hline
& Full-precision & $\mathbb{R}$ & $\mathbb{R}$ & $k^2 i_c  o^2 o_c$ & $0$ & $+,\times$ & 1x\\
\hline

\multirow{5}{*}{\rotatebox{90}{QW$^1$}}
& TTQ & $\mathbb{R}$ & $\left\{-\alpha^{n},0,+\alpha^{p}\right\}$ & $k^2 i_c  o^2 o_c$ & $0$ & $+$ & $\sim 2\times$ \\
 & TWN & $\mathbb{R}$ & $\left\{-\alpha,0,+\alpha\right\}$ & $o^2  o_c$ & $0$ & $+$ & $\sim 2\times$\\
 & BWN & $\mathbb{R}$ & $\left\{-\alpha,+\alpha\right\}$ & $o^2  o_c$ & $0$ & $+$ & $\sim 2\times$\\
 & BC & $\mathbb{R}$ & $\left\{-1,+1\right\}$ & $o^2 o_c$ & $0$ & $+$&$\sim 2\times$ \\
\cline{2-8}
 & BW-CF$\beta$ & $\mathbb{R}$ & $\left\{-\alpha_l,+\alpha_l\right\}$ & $ o^2 o_c / \beta$ & $0$ & $+$& $\sim 2 \times$\\

\hline
\hline

\multirow{6}{*}{\rotatebox{90}{QW+QA$^1$}}
 & BNN/BinaryNet & $\left\{-1,+1\right\}$ & $\left\{-1,+1\right\}$ & $0$ & $8k^2 i_co^2 o_c$ & XOR, bitcount & $64\times$\\
 & XNOR & $\left\{-\gamma,+\gamma\right\}$ & $\left\{-\alpha,+\alpha\right\}$ & $2o^2 o_c $ & $2k^2 i_c  o^2 o_c$ & XOR, bitcount &$58\times$\\
& HORQ & $\left\{-\gamma,+\gamma\right\}\times 2$ & $\left\{-\alpha^{n},0,+\alpha^{p}\right\}$ & $4 o^2 o_c $ & $4k^2 i_c  o^2 o_c$ & XOR, bitcount & $29\times$\\
 & DoReFa & $\left\{0,1\right\} \times 2$ & $\left\{-\alpha_c,+\alpha_c\right\}$ & $o^2 o_c $ & $4k^2 i_c  o^2 o_c$ & AND, bitcount & $30\times$\\
 & TBN & $\left\{-1,0,+1\right\}$ & $\left\{-\alpha,0,+\alpha\right\}$ & $o^2 o_c $ & $3k^2 i_c  o^2 o_c$ & \small AND, XOR, bitcount &$40\times$\\
\cline{2-8}
 & XNOR-CF$\beta$ & $\left\{-1,+1\right\}$ & $\left\{-\alpha_l,+\alpha_l\right\}$ & $ o^2 o_c / \beta$ & $2k^2 i_c  o^2 o_c$ & XOR, bitcount &$122\times$\\

\hline
\hline
\end{tabular}}
\begin{tablenotes}


\item[-] \small About the weight quantification, the DoReFa adopts a constant scaling factor, while other methods adopt filter-wise scaling factor. Our CF method also belongs to the latter except that some of the filters share the same scaling factor.
\item[-] \small 1: QW here is the abbreviation of Quantize Weight and QA here is the abbreviation of Quantize Activation

\end{tablenotes}
\end{table*}

\subsection{Experimental Result on Image Classification}

Here we compare Trained ternary Quantization(TTQ)\cite{zhu2016trained}, Ternary Weight Network(TWN)\cite{1605.04711}, Binary Weight Network(BWN)\cite{rastegari2016xnor} and,
Binary Connect(BC)\cite{courbariaux2015binaryconnect}, with our Binary-Weight based cross filter compression methods(BW-CF). We also compare Binarized Neural Network(BNN)\cite{courbariaux2016binarized}, BinaryNet\cite{tang2017train}, High-Order Residual Quantization(HORQ)\cite{li2017performance}, DoReFa-Net\cite{zhou2016dorefa},
Ternary-Binary Network(TBN)\cite{wan2018tbn} and XNOR\cite{rastegari2016xnor} with our XNOR-Net based cross filter compression methods(XNOR-CF). The $\beta$ or number behinds our CF method refers to the filter compression parameter.

\begin{table}[htb]
\caption{Experiment Result}
\label{resulttable}
\centering
\resizebox{0.48\textwidth}{!}{
\begin{tabular}{c|ccccc}
\hline
\hline
& Dataset & \multicolumn{2}{c}{CIFAR-10} & \multicolumn{2}{c}{ImageNet}\\

& Models & NIN & VGG-7 & AlexNet & ResNet-18 \\

\hline
\hline

 & FP$^1$        & $91.19$   & $92.88$   & $56.6/80.2^{*}$ & $69.3/89.2$ \\

\hline

\multirow{9}{*}{\rotatebox{90}{QW} }
& BC               & -         &$91.73^{*}$&$35.5/61.0^{*}$& - \\
& BWN              & -         &$92.58^{*}$&$56.8/79.4$    & $60.8/83.0$ \\
& TWN              & -         & $92.56$   &$54.5/76.8^{*}$& $61.8/84.2$ \\
& TTQ              & -         & -         &$57.5/79.7$    & $66.6/87.2$ \\
\cline{2-6}
& BW-CF1        & $90.43$   & $93.10$ & $\textbf{54.6/78.0}$ & $60.3/82.6$\\
& BW-CF2        & $90.26$   & $\textbf{93.27}$ & $54.4/77.7$ & $\textbf{60.7/82.9}$\\
& BW-CF4        & $90.52$   & $92.96$ & $54.3/77.8$ & $60.5/82.6$\\
& BW-CF8        & $90.43$   & $93.08$ & $54.2/77.6$ & $60.6/82.9$\\
& BW-CF16       & $\textbf{90.58}$   & $93.03$ & $54.2/77.8$ & $60.4/82.7$\\

\hline

\multirow{11}{*}{\rotatebox{90}{QW+QA}}
& BNN              & -         & $89.85$   & $27.9/50.4$   & -         \\
& BinaryNet        & -         & -         &$46.6/71.1^{*}$& -         \\
& HORQ             & -         &$91.18^{*}$& -             &$55.9/78.9^{*}$\\
& DoReFa$^2$   & -         & -         & $40.1/$-      & -         \\
& TBN              & -         & $90.85$   & $49.7/74.2$   & $55.6/79.0$ \\
& XNOR             & -         &$90.02^{*}$& $44.1/69.2$   & $51.2/73.2$ \\
\cline{2-6}
&XNOR-CF1          & $84.63$   & $91.27$ & $43.7/68.7$ & $49.5/73.3$\\
&XNOR-CF2          & $\textbf{85.43}$   & $91.30$ & $43.5/68.7$ & $49.0/73.2$\\
&XNOR-CF4          & $85.01$   & $91.36$ & $43.7/68.4$ & $\textbf{49.9/74.0}$\\
&XNOR-CF8          & $84.90$   & $\textbf{91.58}$ & $43.5/68.4$ & $49.0/73.2$\\
&XNOR-CF16         & $84.86$   & $91.42$ & $\textbf{43.7/68.7}$ & $49.7/74.1$\\

\hline
\hline
\end{tabular}}
\begin{tablenotes}
\item[-] \small "*" indicates that the results are reproduced by third parties, mainly from the group of TBN\cite{wan2018tbn}. "-" indicates that neither original papers or third party provide the results.
\item[-] \small 1: FP refers to full-precision implementation
\item[-] \small 2: DoReFa-Net is set to 1-bit weight, 1-bit activation, 32-bit gradient for fair comparison.
\end{tablenotes}
\end{table}

\begin{figure}[htbp]
    \centering
    \subcaptionbox{AlexNet Top-1 Accuracy\label{subfig:alex}}[0.23\textwidth]
    {
        \includegraphics[width=.23\textwidth]{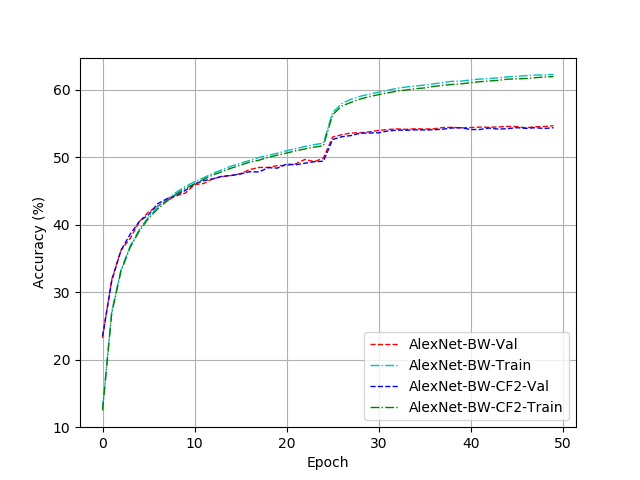}
        \includegraphics[width=.23\textwidth]{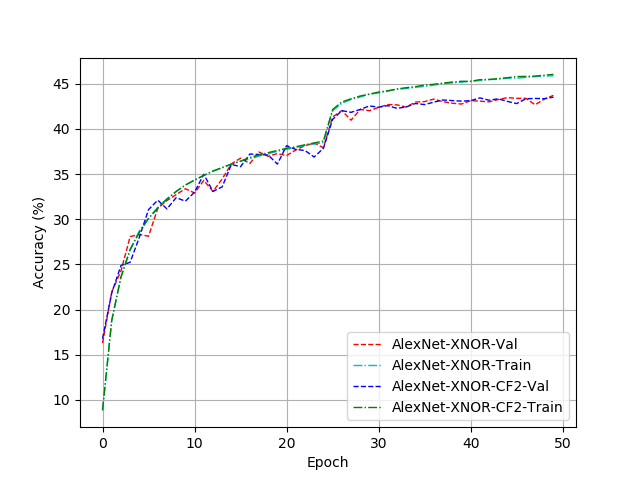}
    }\quad
    \subcaptionbox{ResNet Top-1 Accuracy\label{subfig:res}}[0.23\textwidth]
    {
        \includegraphics[width=.23\textwidth]{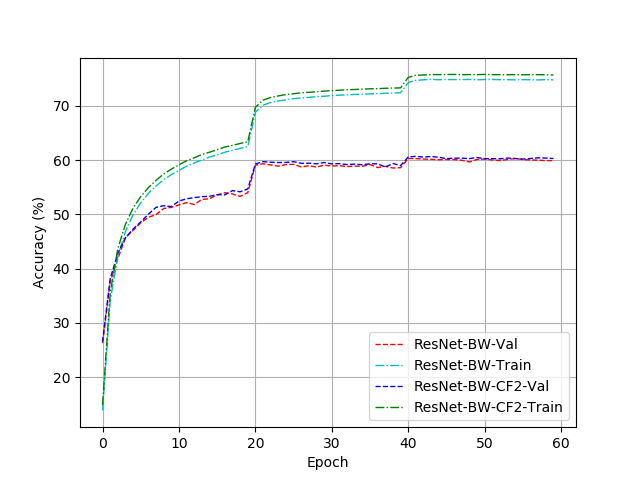}
        \includegraphics[width=.23\textwidth]{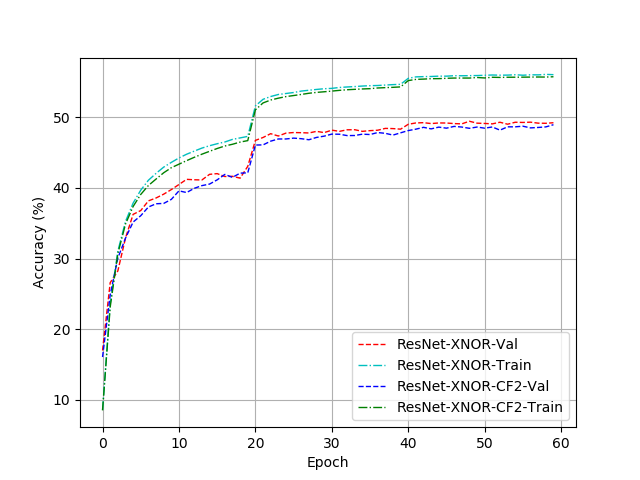}
    }\quad
    \caption{Fig \ref{subfig:alex} compares Top-1 accuracy between BW and BW-CF2, XNOR and XNOR-CF2 on AlexNet, Fig \ref{subfig:res} compares Top-1 accuracy between BW and BW-CF2, XNOR and XNOR-CF2 on ResNet-18}
\end{figure}

\subsubsection{Results on CIFAR-10 with NIN}
The CIFAR-10\cite{krizhevsky2009learning} is a well-known dataset for image classification. It is composed of 60000 images with 32 times 32 pixels from 10 categories. The training dataset contains 50000 images while the test dataset contains 10000 images.

The Network in Network\cite{lin2013network} structure we used is "192-C5 + 160-C1 + 96-C1 + MP2 + 192-C5 + 2x(192-C1) + AP2 + 192-C3 + 192-C1 + 10-C1 + AP1". Here C1, C3, C5 corresponds to 1x1, 3x3 and 5x5 convolution block. MP2 and AP2 is max-pooling/average-pooling layer with kernel size 3, padding 1 and stride 2. AP1 is average-pooling with kernel size 8 and stride 1. No data augmentation or pre-processing is adopted in NIN. The learning rate is set to 0.01 at initial stage and is downscaled by 10 at epoch 120, 200, 240 and 280 with mini-batch size 128. The network is trained for 320 epochs. The initial convolution layer and final convolution layer are kept to full-precision. Other layers are quantified based on our cross-filter methods with different filter compression factor $\beta$.

We evaluate the CIFAR-10\cite{krizhevsky2009learning} dataset with Network in Network\cite{lin2013network} structure. We report the best validation accuracy in Table \ref{resulttable}. Notes that we implement our cross-filter compression method based on both Binary Weight, which only quantify weights(QW), and XNOR, which quantify weights as well as activation(QWQA)\cite{rastegari2016xnor}. Given the fact that many of the methods selected for comparison do not report their result from original papers, we can not see the difference. For QW case, the CF16 model has the highest accuracy. While for QWQA case, the CF1 model has the highest accuracy. We can observe that our methods achieve high performance on CIFAR-10 dataset. And compared to the Binary Weight (BW-CF1) and XNOR(XNOR-CF1), our method achieves the similar accuracy with $\beta$ times reduction in multiplication operations.

\subsubsection{Results on CIFAR-10 with VGG7}
VGG7\cite{simonyan2014very} model is also trained on CIFAR-10 dataset for comparison with other methods. 

The training result can be seen from Table \ref{resulttable}. Notice that we use BW-CF1 to represent BWN because they are the same essentially. In the BW case, CF16 exceeds CF1(BW) by 0.3\% and full precision by nearly 1\%. It also outperforms BC and TWN methods. In XNOR based case, CF8 exceeds XNOR(CF1) by 0.13\% and outperforms all other compression methods, such as HORQ and TBN which require more computational resource than our XNOR-CF8 method.

\subsubsection{Results on ImageNet with AlexNet}
ILSVRC2012\cite{imagenet_cvpr09} image classification dataset contains 1k categories with 1.2M natural images for training and 50K images for validation. 

We implemented our cross-filter compression method and other methods upon AlexNet\cite{krizhevsky2012imagenet} and evaluated them on ILSVRC2012 Dataset. The evaluation result is reported using top-1 and top-5 accuracy. The model structure in this part is "96C11 + 256C5 + 2x(384C3) + 256C3 + MP2 + 2x(L4096) + L1000". Here C11, C3, C5 corresponds to 11x11, 3x3 and 5x5 convolution block. MP2 is max-pooling layer with kernel size 3 and stride 2. L4096 and L1000 refer to linear layer with 4096 and 1000 neurons. The network is trained with batch size 256, epoch 50. The learning rate is 0.001 at initial stage and reduced by 10x at epoch 25. The images are resized to $227\times227$ before fed to the network. Weight Decay is set to $10^{-5}$ and Adam optimizer is used. The first convolution layer and the last linear layer are kept to full-precision, while other layers are quantified accordingly.

From Table \ref{resulttable}, it can be observed that in QW case, our methods outperform BC and is competitive to BWN and TWN with higher compression ratio. TTQ surpass our cross-filter methods by 3\% in Top-1 accuracy. In QWQA case, our method exceeds DoReFa-Net, BNN and achieves similar result as XNOR. The performance of XNOR-CF is worse than TBN and BinaryNet since more aggressive quantification method is implemented on both activation and weights. The training process can also be seen from Fig \ref{subfig:alex}.

\subsubsection{Results on ImageNet with ResNet18}
We also evaluate our cross-filter compression method upon ResNet-18\cite{he2015resnet}. The network is trained with batch size 128, epoch 60. The learning rate is 0.001 at initial stage and reduced by 10x every 20 epochs. The first convolution layer, the shortcut and the last linear layer are kept to full-precision.

The final results are reported in Table \ref{resulttable} and Fig \ref{subfig:res}. It can be seen that our BW-CF and XNOR-CF are competitive to Binary Weight Net and XNOR Net respectively. The BW-CF2 and XNOR-CF4 are the best ones and outperform the filter-wise compression method by 0.4\%. However, the selection of cross-filter scaling factor $\beta$ does not seems to have massive impact on the final result.

\section{Discussion}
\subsection{Similarity between scaling factors and selection of layer compression}
In this section, we will discuss the similarity between scaling factors $\alpha$ for different layers and the overall situation. We wish to explore whether this cross-filter compression method is suitable for different tasks and different networks. Intuitively, the more similar scaling factors $\alpha$ for spatial-adjacent filters are, the less information was lost during cross-filter compression. The successful implementation of our cross-filter compression method may lay on the similarity between scaling factors in the same layer. Hence, we intend to further explore the distribution of scaling factors of each convolution layer. 

\begin{figure}[htb]
\centering
\includegraphics[width=0.4\textwidth]{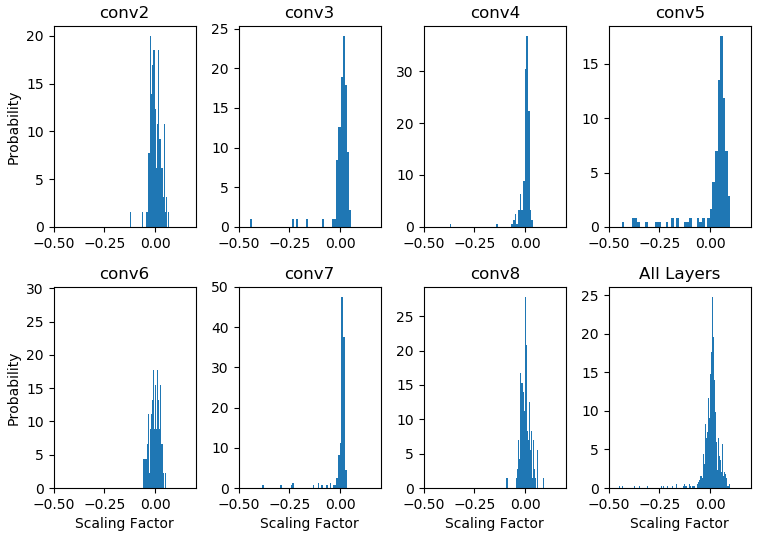}
\caption{\small this is the histogram figure over the distributions of scaling factor in different convolution layers and all layers. All distributions are normalized to zero.} 
\label{fig:diss}
\end{figure}

We import a model trained with binary weight network based on Network in Network over CIFAR-10 dataset. A histogram based on the scaling factors of each convolution layer can also be seen in Fig.\ref{fig:diss}. It can be observed that the distribution of scaling factor $\alpha$ is highly concentrated.

However, from Fig.\ref{fig:diss}, we can observe that distribution of scaling factors in different layers varies from each other. conv4, conv7 and conv8 layers show more consistent distribution than conv3 and conv5 layers. This feature can be used when selecting the layers to be compressed by our cross-filter compression method.

\subsection{Relieving the combination dilemma between uniform quantization and compact network}
The motivation of our idea originates from the potential conflict between the compact network and uniform quantization. In this section, we will shows that our cross-filter compression method has relieved such problem.
Suppose we adopt our cross-filter compression methods, then the number of full-precision MAC operations is reduced by $\beta$ times for both $5\times5$ filter and two $3\times3$ filters, which we used as example in section \ref{dilemma section}. The equation \ref{dilemma} can be rewritten as:
\begin{equation}
    \frac{1}{L}25i_c^2N_o + \frac{1}{\beta}\gamma i_cN_o \leq \frac{1}{L}18i_c^2N_o + \frac{2}{\beta}\gamma i_cN_o 
\end{equation}
and can be further simplified as:
\begin{equation}
    i_c \geq \frac{\gamma}{7\beta}L
\end{equation}
If we adopt $\beta=16$, $\gamma=1.91$ and $L=512$, the boundary becomes 8.7, which can be neglected in neural network. Hence, we can conclude that our cross-filter compression method overcome the potential drawback between compact network design and uniform quantization in current situation.

\subsection{Quantizing first and last layer}
In most cases, the first and last layer of the convolution neural network are omitted when applying compression and acceleration method to the network. It is commonly believed that the first and last layer, which deals with the input and output directly, requires floating point precision in order to maintain model accuracy and capability. Quantifying these two layers usually leads to dramatic decrease in accuracy.

This section aims to explore the influence of quantifying the first and last layer with our proposed cross-filter compression method. With the increase of filter compression number $\beta$, the number of multiplication decreases massively. However, multiplication operations in the first and last layer also gradually become the bottleneck of our method.

We implement BW-CF2 and XNOR-CF2 methods upon Network in network. The training configuration is the same as it is in previous experiment. The final result is shown in Table \ref{Quan-NIN}.
We can observe that quantizing the first layer results in severe accuracy lost. In BW-CF2, the accuracy drop for 2.4\%, while in XNOR-CF2, the accuracy reduction even increase to 8\%. 
For the last layer, the cost of quantization is much small than that of the first layer. In BW-CF2 case, the accuracy even increase by 0.30\%. In XNOR-CF2, the accuracy drop is 1.50\%.

\begin{table}[htbp]
\caption{Quantifying first and last layer in NIN} 
\label{Quan-NIN}
\centering 
\begin{tabular}{c c c } 
\specialrule{0em}{1.5pt}{1.5pt}
\hline\hline 
\specialrule{0em}{1.5pt}{1.5pt}
Quantification & BW-CF2(Red.) & XNOR-CF2(Red.) \\ [0.5ex] 
\hline 
\specialrule{0em}{1.5pt}{1.5pt}
NIN        & 90.15 & 85.95  \\
NIN+first  & 87.79(2.36) & 77.92(8.03)  \\ 
NIN+last   & 90.45(-0.30) & 84.55(1.50)  \\
\hline 
\end{tabular}
\end{table}

\section{Conclusion}
In this paper, we proposed a cross-filter compression method that shares numerically similar scaling factors cross filters to accelerate convolutional neural networks (CNNs). Our technique enables the application of uniform quantization approach to a compact network with small kernels such that the two conflicting methods can be combined for CNN acceleration. The proposed technique can be widely applied to various CNN inference applications and practically improve the real-time performance for edge intelligence devices. Integrated upon XNOR-Net, it can speed up convolution layer by up to $122\times$ and save $\sim32\times$ model memory, which outperforms the state-of-the-art compression methods. This method breaks the filter-wise limitation compared to other quantization methods and has the potential to be extended to all quantification method with consistent scaling factor for the same filter. It produces result with up tp 0.3\% accuracy gain compared to original filter-wise quantization methods. We also prove that the optimal compression result can be obtained not only from full-precision models but also from models quantizated with filter-wise methods. This method was evaluated on different network: NIN\cite{lin2013network}, VGG\cite{simonyan2014very}, AlexNet\cite{krizhevsky2012imagenet} and ResNet\cite{he2015resnet} upon various datasets: cifar-10\cite{krizhevsky2009learning} and ImageNet\cite{imagenet_cvpr09}.

\bibliographystyle{named}
\bibliography{ijcai19}

\end{document}